\documentclass[runningheads]{llncs}
\usepackage{graphicx}
\usepackage{amsthm}
\usepackage{mathtools}
\usepackage{amsmath}
\usepackage{amssymb}

\usepackage{multirow}
\usepackage{soul}
\usepackage[utf8]{inputenc}
\usepackage[small]{caption}
\usepackage{booktabs}
\usepackage{algorithm}
\usepackage{algorithmic}
\usepackage{color}
\usepackage{dsfont}
\usepackage[normalem]{ulem}
\usepackage{makecell}
\usepackage[table,xcdraw]{xcolor}
\usepackage[english]{babel} 
\usepackage{blindtext}

\DeclareMathOperator*{\argmax}{arg\,max}

\DeclareMathOperator{\define}{\coloneqq}
\DeclareMathOperator*{\Real}{Re}
\DeclareMathOperator*{\Ima}{Im}

\begin{document}
\title{Learning the Propagation of Worms in Wireless Sensor Networks}
%
%
\author{Yifan Wang\orcidID{0000-0002-6312-4663} \and
Siqi Wang\orcidID{0000-0002-0089-1005} \and
Guangmo Tong\orcidID{0000-0003-3247-4019}}
\authorrunning{Y. Wang et al.}
%
\institute{University of Delaware, Newark DE 19711, USA\\
\email{\{yifanw,wsqbit,amotong\}@udel.edu}}
\maketitle              
\begin{abstract}
Wireless sensor networks (WSNs) are composed of spatially distributed sensors and are considered vulnerable to attacks by worms and their variants. Due to the distinct strategies of worms propagation, the dynamic behavior varies depending on the different features of the sensors. Modeling the spread of worms can help us understand the worm attack behaviors and analyze the propagation procedure. In this paper, we design a communication model under various worms. We aim to learn our proposed model to analytically derive the dynamics of competitive worms propagation. We develop a new searching space combined with complex neural network models. Furthermore, the experiment results verified our analysis and demonstrated the performance of our proposed learning algorithms.

\end{abstract}
\section{Introduction}

Due to the recent advances in wireless and micro-mechanical sensor technologies, wireless sensor networks (WSNs) have been widely used in various real-time applications, for example, target tracking, biological detection, health care monitoring, environmental sensing and etc. \cite{khanh2016dynamics,regin2021fault,ojha2021improved,nain2021localization,krishnamachari2005networking,galluccio2019impact}. WSNs are composed of the spatially distributed sensors that are low-cost, small-sized, low-energy, and multi-functional \cite{khanh2016dynamics,wang2013modeling,gulati2021review}. Each sensor is designed to sense and collect environmental information. It has the ability to carry out a local computation to obtain the required data and will transfer the required data to a higher-level host \cite{akyildiz2002wireless,wang2013modeling}. Besides, the communication paradigm of WSNs is based on the broadcasting such that each sensor can communicate with its neighbors directly \cite{wang2010eisirs,akyildiz2002wireless}.

The computer worms are the programs that can be disseminated across the network by self-replicating to exploit the security or policy flaws \cite{weaver2003taxonomy,rajesh2015survey}. Although many advantages of the sensors are proposed, WSNs are vulnerable to attacks by worms and variants. One reason is that sensor network protocols primarily concentrate on power conservation, and each sensor is usually assigned to a limited resource. Therefore it generally has a weak defense mechanism \cite{mishra2013mathematical,srivastava2016stability}. Besides, since the spread of worms usually relies on the information involved in the infected sensors and WSNs often embed an efficient propagation mechanism, the worms can achieve a rapid propagation speed \cite{awasthi2020study,khanh2016dynamics,achar2022dynamics}. Furthermore, some worms will upgrade and be more complicated in order to propagate successfully, which will result in a widespread destruction \cite{wang2013modeling}. Therefore, worms and variants become one of the main threats to network security.

In order to control the spreading of worms, people seek to understand the dynamic behaviors of worms. In practice, worms utilize different propagation strategies. For example, scan-based techniques will identify the vulnerable sensors based on the IP address \cite{wang2011microcosmic,zou2005routing}. Topology-based techniques allow worms to spread through the topology neighbors \cite{wang2013modeling}. Many studies have modeled these propagation procedures mathematically. Most researchers proposed propagation models based on the classic epidemic models, for example, SIR and SEIR models \cite{kephart1993computers,han2010dynamical}. In recent works, extensions to the epidemic models have been proposed: Some studies considered the hybrid dynamic of worms and antivirus propagation \cite{abdel2021unification,behal2022dynamics,haghighi2016race}. Some works investigated the virus behavior containing the stability analysis of WSNs \cite{zarin2022deterministic,hu2015stability,qin2015analysis}.                                        

Considering a WSN composed of a number of sensors, the transmission rates for different worms vary based on the distinct mechanism embedded in the sensors. That means that for different worms and sensors, the dynamic behavior will vary \cite{chakrabarti2007information,yao2019propagation}. Meanwhile, both scan-based and topology-based propagation strategies select the targets based on the features related to the sensors \cite{wang2013modeling,nwokoye2022epidemic}. Since there is a basic similarity between worm propagation through wireless devices and the information diffusion process over the regular network, from this standpoint, we propose a discrete communication model based on the competitive diffusion models under multiple worms. We seek to learn our proposed model to derive the dynamics of competitive worms propagation analytically. Our contributions can be summarized as follows.
\begin{itemize}
    \item \textbf{Propagation Model design.} We design a communication model that simulates the propagation of the competitive worm. Our model includes the special case with the hybrid dynamics with virus and patch, which provides a new aspect to the antivirus protection design.
    \item \textbf{Searching Space Analysis.} We develop a new searching space combined with the complex-valued neural network. To our best knowledge, this provides a starting point for analyzing the worms propagation model. Besides, a back-propagation-based learning algorithm for the complex-valued neural network has been proposed.
    \item \textbf{Empirical Studies.} We provide experimental validation of our model compared with some baseline methods.
\end{itemize}

The paper is organized as follows. In the next section, we introduce some related works of the different propagation models. Section 3 illustrates the structure of the communication model and formulates the learning problem. Section 4 describes the learning framework, including the searching space design and the learning algorithm. Experiments are given in section 5.

\section{Related Work}
A large number of models have been proposed to analyze the dynamics of worm propagation in WSNs based on epidemic models. Kephart et al. first proposed the idea of formulating the worm's propagation based on simple epidemic models. They assumed each sensor or host stays in one of two states: susceptible or infectious. And the infected sensor will not change its infectious state \cite{kephart1993computers}. In recent works, Abdel et al. proposed an extension SIR model by considering antivirus dynamics that add the antidotal state to each sensor \cite{abdel2021unification}. And the authors present a unified study of the epidemic-antivirus nonlinear dynamical system. The work in \cite{achar2022dynamics} designed a fractional-order compartmental epidemic model by accounting for the global stability analysis. Several studies added the quarantined constraints for each sensor to the epidemic models \cite{khanh2016dynamics,mishra2014defending,mishra2014quarantine}.  

For multiple worms propagation, authors in \cite{wang2010eisirs} proposed a model that supports the sleep and work interleaving schedule policy for the sensor nodes. They assigned different infect probabilities to each worm when the infected nodes attempted to infect other nodes. However, they did not specify the relationship among these worms, and it is hard to track the dynamic of different worms. \cite{nwokoye2020evaluating} developed a SEjIjR-V Epidemic Model with differential exposure, infectivity, and communication Range. Besides, they generate the epidemic thresholds for each sensor corresponding to different worms. In our work, we formulate a worms propagation model based on the works in \cite{wang2022learnability}. And inspired by them, we develop an explicitly complex-valued neural network to exactly simulate the worms' propagation process.

\section{Preliminaries}
In this section, we present the communication model and problem settings.

\subsection{Communication Model}

We consider the wireless network as a directed graph $G=(V,E)$, where $V$ and $E$ denote the sensors and the communication links, respectively. Let $|V|=N$ and $|E|=M$. For each node $v$, its in-neighbors is denoted by $N_v\subseteq V$. It is assumed that $K$ different worms are spread over the network, and each worm $k\in [K]$ propagates from a set of sensors $S^k \in V$. For each node $v\in V$, there is a set of threshold $\Theta_v = \{\theta_v^1,...,\theta_v^K\}$ that represents its vulnerabilities to different worms, where $\theta\in [0, \infty)$. For each edge $(u,v)\in E$, there is a set of weights $W_{(u,v)} = \{w_{(u,v)}^1, ..., w_{(u,v)}^K\}$ indicates its levels of transmission with different worms, where $w\in [0, \infty)$. We assume that a node can only be infected by one worm and suppose that the spread of worms continuous step by step based on the following rules.

\begin{itemize}
    \item Initialization. For nodes $v\in S^k$, they are $k$-infected; Otherwise, the nodes stay innocent.
    \item Step $t$. There are two stages during each step. Let $S^k_{t-1}\subseteq V$ be the $k$-infected nodes at the end of step $t-1$.
    \begin{itemize}
        \item Stage 1. For each innocent node $v$, it can be possibly infected by the worms in set $P_v^t \subseteq [K]$, where
            \begin{equation}
                P_v^t = \Big\{k\Big|\sum_{u\in N_v \cap S^k_{t-1}}w_{(u,v)}^k \geq \theta_v^k\Big\}.
            \end{equation}
        \item Stage 2. Node $v$ will finally be infected by worm $k^*$ such that
            \begin{equation}
                k^* = \argmax_{k\in P_v^t} \sum_{u\in N_v \cap S^k_{t-1}}w_{(u,v)}^k
            \end{equation}
            Once a node is infected, its status will not change during the propagation.  For tie breaking, each node is more vulnerable with the worm with a large index. 
    \end{itemize}
    \item Termination. The propagation process ends when no nodes can be infected. We can observe that $t\leq N$.
  
\end{itemize}

\subsection{Problem Settings}
We begin by presenting some notations that are used in the rest of the paper.

\textbf{Notations.} For a complex number $a+ib$, $a$ is the real part and $b$ is the imaginary part, where $a, b \in \mathbb{R}$. For convenience, we use $(a, b)$ to simplify the notation. During the propagation process, for a step $t\in [N]$, we denoted the status of each node $v\in V$ by a complex-valued vector $I_t(v) \in \mathbb{C}^K$: If $v$ is $k$-infection, then the $k_{th}$ element in $I_t(v)$ is $1+i0$ and the rest elements are $0+i0$; If $v$ is a innocent node, then all the elements in $I_t(v)$ are $0+i0$. The status over all nodes, named all-infection status, is then formulated as a matrix $\mathcal{I}_t \in \mathbb{C}^{K\times N}$  by concatenating $I_t$ for each node, where $\mathcal{I}_{t,v}=I_t(v)$. Furthermore, to simplify, we assume that $K = 2^P$ where $p\in \mathbb{Z}^+$.

We are interested in learning propagation functions given the collected propagation data. 
\begin{definition}
Given a network graph and a communication model, the propagation function $F: \mathbb{C}^{K\times N} \to \mathbb{C}^{K\times N}$ maps from initial all-infection status $\mathcal{I}_0$ to the final status $\mathcal{I}_N$.
\end{definition}

\begin{definition}
The propagation data set $D$ is a set of initial-final status pairs with size $Q\in \mathbb{Z}^+$, where
\begin{equation}
    D = \{(\mathcal{I}_0, \mathcal{I}_N)^i\}_{i=1}^{Q}.
\end{equation}
\end{definition}

Given a new initial-final status pair $(\mathcal{I}_0, \mathcal{I}_N)$, let $\hat{\mathcal{I}}_N$ be the predicted final all-infection status, we use $l$ to measuring the loss. 
\begin{equation}
    l(\hat{\mathcal{I}}_N, \mathcal{I}_N) = \frac{1}{N}|\{v:\hat{\mathcal{I}}_{N,v} \neq \mathcal{I}_{N,v}, v\in V\}|
\end{equation}

The problem can be formally defined as follows.
\begin{problem}\label{pro}
Given the graph structure $G$ and a set of propagation data $D$ draw from some distribution $\mathcal{D}$, we wish to find a propagation function $F: \mathbb{C}^{K\times N} \to \mathbb{C}^{K\times N}$, that minimizes the generalization error $L$, where 
\begin{equation}\label{equ:generalization_error}
    L(F) = \mathbb{E}_{(\mathcal{I}_0, \mathcal{I}_N) \sim \mathcal{D}}l(F(\mathcal{I}_0), \mathcal{I}_N)
\end{equation}
\end{problem}

\section{A Learning Framework}
In this section, for the communication model, we first design a searching space that composed by the complex neural network. Then the learning algorithm is proposed given such searching space. 
\subsection{Hypothesis Space Design}
In this section, we construct the hypothesis space for problem \ref{pro} by simulating the propagation process as a finite complex-valued Neural Network (CVNNs) from local to global.  Assume that for the complex-valued neuron $n$ at layer $l+1$, it is linked with $m$ numbers of neurons in the previous layer, then the output of neuron $n$ is described as follows:
\begin{equation}
    H^{l+1}_n = \sigma^{l+1}_n(\sum_{i=1}^m W^{l}_{in}H^{l}_n + T_n^l)
\end{equation}

Here $W^{l}_{in}$ denotes the weight connecting the neuron $n$ and the neuron $i$ from the previous layer, $\sigma^{l+1}_n$ denotes the activation function on neuron $n$, $T_n^l$ is the bias term.

\subsubsection{Local Propagation}

\begin{figure*}[!t]
    \centering
    \includegraphics[width = \textwidth]{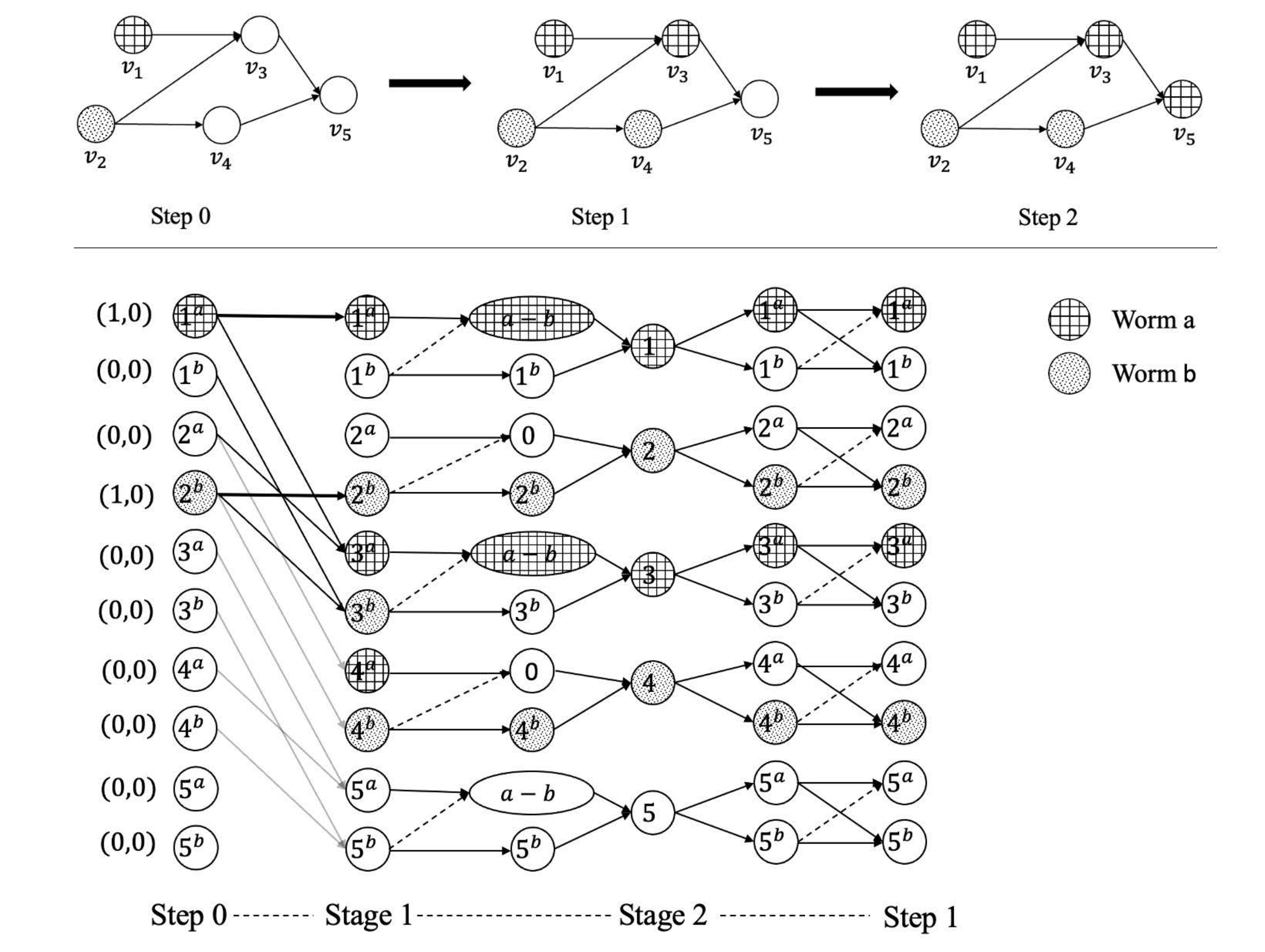}    
    \caption{\textbf{Local propagation structure.} Suppose that two worms are disseminated in the network. The top half of the figure shows the whole propagation process across a small WSN. The bottom half shows the structure of our proposed complex neural network during step 1. Each stages are simulated by several layers in the network. And three additional layers are added to control the format of the output.}
    \label{fig:local}
    \end{figure*}

For the local propagation, it simulates the diffusion process during every step $t$. It starts by receiving the all-infection status $\mathcal{I}_t \in \mathbb{C}^{K\times N}$ at the end of step $t$. Figure \ref{fig:local} illustrates an example of local propagation process.

\textbf{Stage 1.} We aim to find out all the possible worms that can infect the each innocent nodes. During this stage, each worm diffuses separately and follows a linear threshold process. Therefore, we wish the links between the layers represents the topology in the WSNs and the activation function can be simply defined as the threshold functions. In addition, the communication model indicates that once a node is affected, it will never change the status. Therefore, for each node, we add one link towards it self at the next layer with a relatively large value. The propagation can be simulated by
        \begin{align}
        \label{eq: H1_H2}
        \mathbf{H}^{2} = \sigma^{1}(\mathbf{W}^{1}\mathbf{H}^{1}+\mathbf{T}^1),  
        \end{align}
    
    where for $i, j \in [N]$ and $k, l \in [K]$,  we have
    \begin{align*}
    \mathbf{W}^{1}_{i\cdot K + k, j\cdot K + l } \define 
        \begin{cases}
          (w_{(i,j)}^k, 0) & \text{if $(i, j)\in E$ and $k=l$} \\
          1000 & \text{if $i=j$ and $k=l$} \\
          (0, 0) & \text{otherwise} 
        \end{cases}
     \end{align*}
    
    \begin{equation}
        \sigma ^{1}_{j\cdot K + l}(x) \define 
        \begin{cases} 
          x & \text{if $\Real(x)\geq \theta_j^l$} \\
          (0, 0) & \text{Otherwise} 
        \end{cases}.
        \end{equation}
    
    \begin{equation}
        T ^{1}_{j\cdot K + l}(x) \define (0, l)
        \end{equation}

\textbf{Stage 2.} With the input $\mathbf{H}^2$ from phase 1, this stage determines the final infect status of each node. That is for a node $i \in V$, we want to find out the largest value among the sub-array 
    \begin{equation}\label{equ:subarray}
    \Big[\Real(\mathbf{H}^2_{i + 1}), \Real(\mathbf{H}^2_{i + 2}),..., \Real(\mathbf{H}^2_{i+K})\Big].
    \end{equation}
Intuitively, finding the largest value among equation \ref{equ:subarray} can be implemented through a hierarchy structure of pairwise comparisons.  Starting from the sub-array, each hierarchy compares every two neighboring elements and passing the larger value to the next level, which leads to a $P$ hierarchy levels structure.
    \begin{figure*}[!t]
    \centering
    \includegraphics[width = \textwidth]{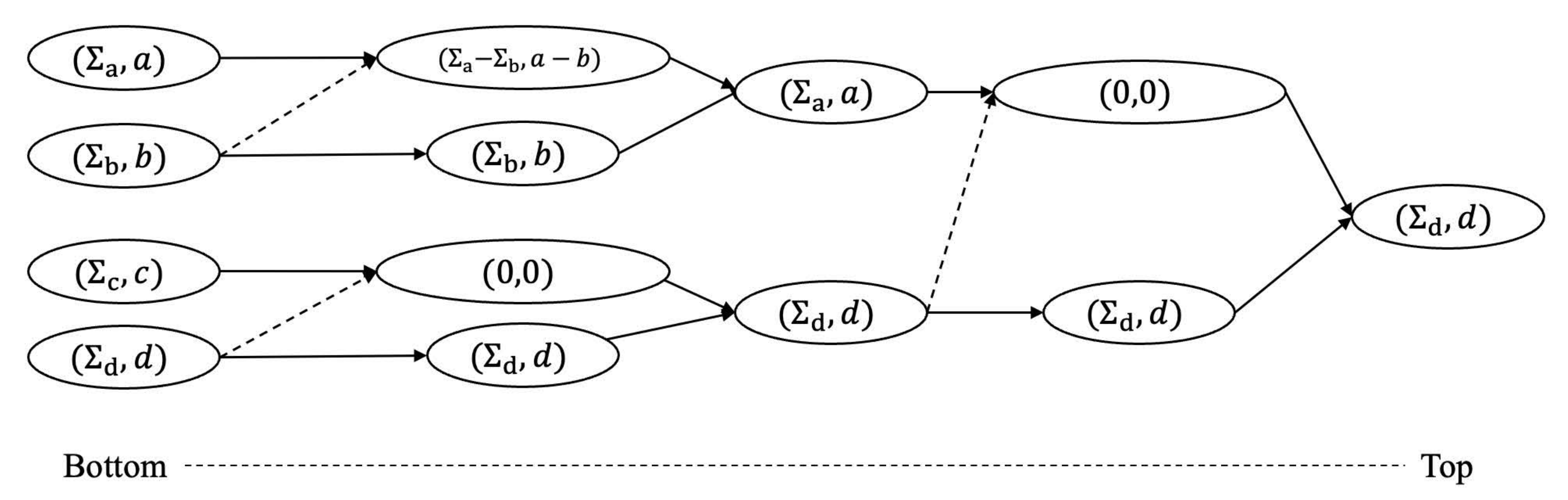}    
    \caption{\textbf{Hierarchy structure of pairwise comparisons.} Suppose that there are four different worms: $a, b, c$ and $d$. There are two comparison process with two layers. The nodes in the bottom layer represent the sub-array \ref{equ:subarray}. All the weights on the black solid links are $1+i0$ and the weights on the dash links are $-1+i0$.  During each comparison process in each two nodes, the first layer implements the activation function in equation \ref{eq:comparison_function} such that: if the prior one has the larger value of real part, the node returns the difference both in real and imaginary part; Otherwise, it will return $(0, 0)$. Therefore, by combining the second layer, the output is exactly the larger value between two nodes.}
    \label{fig:comparison}
    \end{figure*}
    From bottom to top, for node $v\in [N]$, in each hierarchy level $p\in [P]$, the size of input elements $I_p = \frac{K}{2^{p-1}}$ and the size of output elements $O_p = \frac{K}{2^{p}}$. Each level can be decoded by two layers: $l^{c_1}$ and $l^{c_2}$. 
    
    In $l^{c_1}$, for $i, j \in [I_p]$,
        \begin{align*}
        \mathbf{W}^{c_1}_{i, j} \define
        \begin{cases}
            (1, 0) & \text{if $i=j$}\\
            (0, -1) & \text{if $j=i-1$ for $j = 2n+1$ where $n\in \{0, 1, 2,..., I_p-1\}$}\\
            (0, 0) & \text{otherwise} 
         \end{cases}
        \end{align*}
    $\sigma^{l_{c1}}$ is defined as follows:
        
        \begin{equation}\label{eq:comparison_function}
        \sigma^{l_{c_1}}(x) \define
        \begin{cases}
            (\Ima(x), \Real(x)) & \text{if $\Ima(x)\geq 0$}\\
            (0, 0) & \text{otherwise} 
        \end{cases}
        \end{equation}
    
    In $l^{c_2}$, for $i \in [I_p]$ and $j\in [O_p]$, 
        \begin{flalign*}
        &\mathbf{W}^{l_{c_2}}_{i, j} \define 
            \begin{cases}
            (1, 0) & \text{if $i=2\cdot j$ or $i =2\cdot j -1 $}\\
            (0, 0) & \text{otherwise} 
            \end{cases}
        \end{flalign*}
    
    $\sigma^{l_{c_2}}$ is defined as follows.
        \begin{align*}
        \sigma ^{l_{c_2}}(x) \define x
        \end{align*}
    
    By repeatedly using these two layers $P$ times, the output of stage 2 is $\mathbf{H}^{2P+2} \in \mathbb{C}^N$. For $i\in [N]$ and $k\in [K]$, each node in $\mathbf{H}^{2P+2}$ holds the value as follows,

    \begin{equation}\label{equ:StageTwoOutput}
    \mathbf{H}^{2P+2}_i= \Big(\max_{k\in K}\{\Real(\mathbf{H}^2_{i + k})\}, \argmax_{k\in K}\{\Real(\mathbf{H}^2_{i + k})\} \Big),
    \end{equation}
    
    where the real part for each node $i$ indicates the largest incoming weights among all the worms, and the imaginary part is the corresponding worm index.
    
\begin{figure*}[!t]
\centering
\includegraphics[width = 0.7\textwidth]{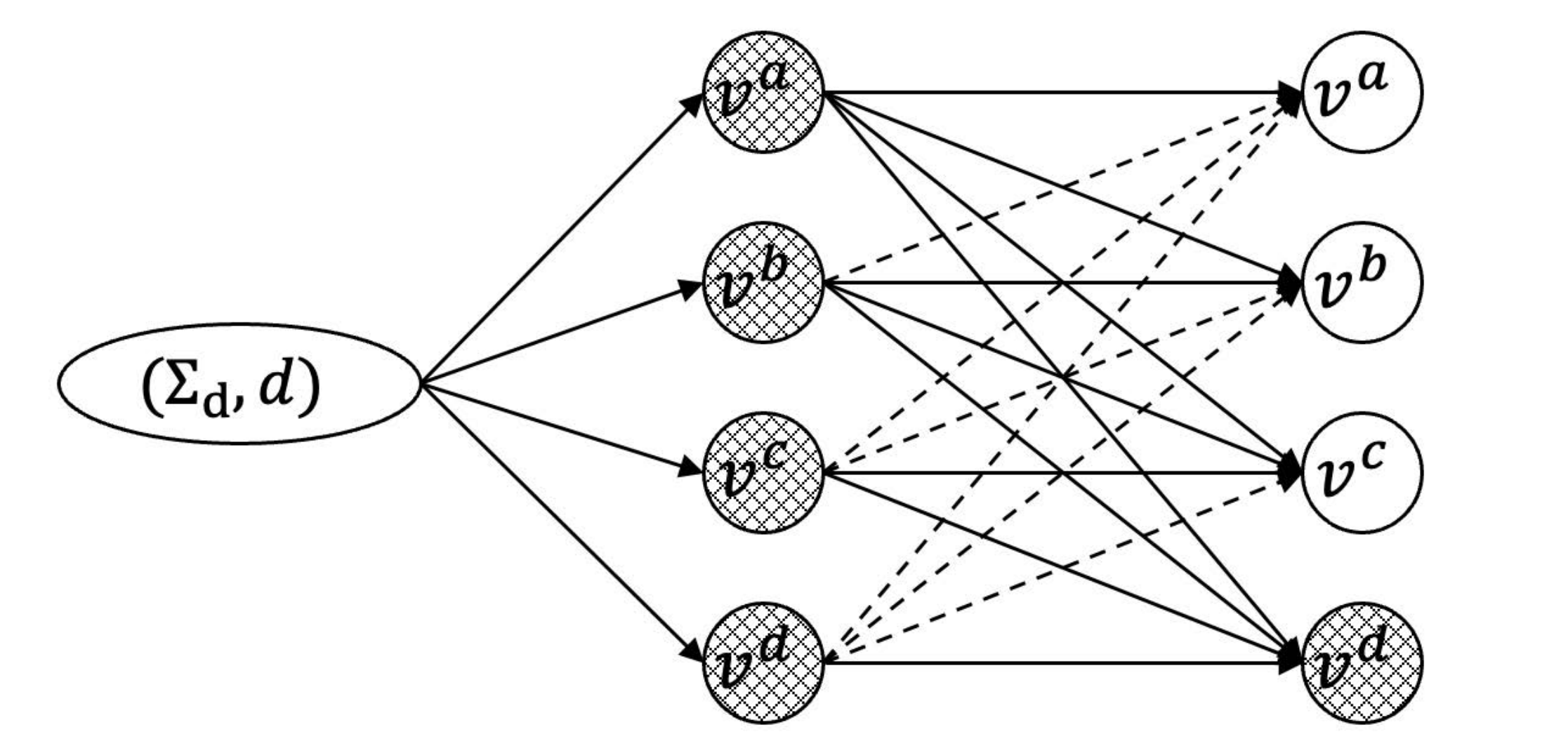}    
\caption{\textbf{Structure of the format adjustment.} For node $v\in V$, suppose that there are four different worms: $a, b, c$ and $d$. All the weights on the black solid links are $1+i0$ and the weights on the dash links are $-1+i0$. The value on the nodes with shadow is $1+i0$; Otherwise, it is $0+i0$. In addition, the threshold on each node $v^k$ where $k\in\{a, b, c, d\}$ is $k$.}
\label{fig:recovery}
\end{figure*}

\textbf{Format Adjustment.} In order to simulate the global worm propagation process, we need to make the format of the output at the end of local propagation be same as the input, that is $\mathbf{H}^{2Z+3} = \mathcal{I}_{t+1} \in \mathbb{C}^{K\times N}$. Two layers are added. Figure \ref{fig:recovery} shows an example of the structure of the format adjustment. The two layers can be formulated as follows.

    \begin{align}
    \label{eq: final}
    \mathbf{H}^{2P+2+i} = \sigma^{2Z+2+i}(\mathbf{W}^{2Z+2+(i-1)}\mathbf{H}^{2Z+2+(i-1)}),  
    \end{align}
    
where $i=\{1, 2\}$. The first layer can be seen as a transition layer. The number of the output nodes will be recovered as the input, and the real number of each node are set to $0$. For the imaginary part, its value will be equal to $1$ if the node's corresponding worm's index is no large than the real part in equation \ref{equ:StageTwoOutput}. Therefore, it is formulated as follows. For each $i, j \in [N]$, $k\in [K]$,  we have
    \begin{align*}
    \mathbf{W}^{2P+3}_{i\cdot K, j \cdot K + k} \define 
        \begin{cases}
            (1, 0) & \text{if $i=j$}\\
          (0, 0) & \text{otherwise} 
        \end{cases}
    \end{align*}
    
    and       
    \begin{equation}
        \sigma_{j\cdot K+k}^{2P+3}(x) \define 
        \begin{cases} 
          (1, 0) & \text{if $\Ima(x)$ and $\Ima(x)\geq k$} \\
          (0, 0) & \text{Otherwise} 
        \end{cases}.
        \end{equation}

The last layer are designed to extract the all-infection status at the end of local propagation, and that is $\mathbf{H}^{2P+4} = \mathcal{I}_t$
    
    where for each $i, j \in [N]$, $k, l\in [K]$,  we have
    \begin{align*}
    \mathbf{W}^{2P+4}_{i\cdot K+k, j\cdot K+l} \define 
        \begin{cases}
            (1, 0) & \text{if $i=j$ and $l \geq k$}\\
            (-1, 0) & \text{if $i=j$ and $l < k$}\\
            (0, 0) & \text{otherwise} 
        \end{cases}
    \end{align*}
    
    and       
    \begin{equation}
        \sigma_{j\cdot K +l}^{2P+4}(x) \define 
        \begin{cases} 
          (1, 0) & \text{if $\Real(x)>l$} \\
          (0, 0) & \text{Otherwise} 
        \end{cases}.
        \end{equation}
We can then obtain the following lemma.
\begin{lemma}
The local propagation under transmission model can be simulated by a complex neural network with $2P+4$ layers.
\end{lemma}

\subsubsection{From Local to Global}
The output of local propagation complex neural network can be viewed as the input for the next time step. We can repeatedly using the local network to exactly simulated the whole propagation process. We denoted such global complex neural network by $M_C$. Therefore, we can obtain our hypothesis space of problem \ref{pro}.
\begin{definition}
Taking the weights $W_{(u,v)}$ for $(u,v)\in E$ and thresholds $\Theta_v$ for $v\in V$ as the parameters, the hypothesis space $\mathcal{F}$ is the set of all possible $M_C$. 
\end{definition}

\subsection{Learning Algorithm}
Given the hypothesis space $\mathcal{F}$, the propagation function $F$ based on the propagation model is now implemented as complex neural network models. Instead of directly learning the $F$, we seek to learn a complex neural network model from $\mathcal{F}$ that can minimize the generalization error in equation \ref{equ:generalization_error}. Same as the neural network, learning in complex neural networks refers to the process of modifying the parameters involved in the network such that the learning objectives can be achieved. We choose the gradient-based approach with the complex domain back-propagation \cite{bassey2021survey}. During the training, We utilize the complexPytorch package, which is a high-level API of PyTorch \cite{matthes2021learning,trabelsi2018deep}.

\section{Experiments}
In this section, we evaluated our proposed method by experiments. We compare our method with several baseline algorithm on synthetic and real worlds networks. 
\begin{table}[]
\centering
\caption{\textbf{Main results.} Each simulation is implemented with 600 training samples, 400 testing samples and 4 worms.}
\label{tab:main}
\vspace{2mm}
\resizebox{\textwidth}{!}{%
\begin{tabular}{cccccc}
                                 &                 & F1 score                               & Precision                              & Recall                                 & Accuracy                               \\ \hline
\multirow{4}{*}{ER graph}        & Proposed method & 0.469\tiny(6.E-03)                     & 0.419\tiny(6.E-03)                     & 0.629\tiny(6.E-03)                     & 0.727\tiny(3.E-03)                     \\
                                 & RegularNN       & 0.408\tiny(4.E-03)                     & 0.329\tiny(4.E-03)                     & 0.590\tiny(6.E-03)                     & 0.695\tiny(3.E-03)                     \\
                                 & SVM             & 0.184\tiny(0)                          & 0.192\tiny(0)                          & 0.201\tiny(0)                          & 0.498\tiny(1.E-03)                     \\
                                 & Random          & 0.182\tiny(2.E-03)                     & 0.200\tiny(2.E-03)                     & 0.200\tiny(2.E-03)                     & 0.200\tiny(2.E-03)                     \\ \hline
\multirow{4}{*}{Sensors network} & Proposed method & 0.408\tiny(1.E-02)                     & 0.372\tiny(1.E-02)                     & 0.575\tiny(1.E-02)                     & 0.684\tiny(6.E-03)                     \\
                                 & RegularNN       & 0.208\tiny(1.E-02)                     & 0.302\tiny(6.E-02)                     & 0.450\tiny(3.E-02)                     & 0.597\tiny(3.E-03)                     \\
                                 & SVM             & \multicolumn{1}{l}{0.146\tiny(8.E-04)} & \multicolumn{1}{l}{0.117\tiny(3.E-03)} & \multicolumn{1}{l}{0.205\tiny(3.E-04)} & \multicolumn{1}{l}{0.537\tiny(3.E-03)} \\
                                 & Random          & \multicolumn{1}{l}{0.176\tiny(7.E-03)} & \multicolumn{1}{l}{0.200\tiny(4.E-03)} & \multicolumn{1}{l}{0.200\tiny(8.E-03)} & \multicolumn{1}{l}{0.199\tiny(7.E-03)}
\end{tabular}%
}
\end{table}

\subsection{Settings}

\textbf{WSN data.} We adopt two types of graph structures to evaluate our proposed method: one synthetic graph and a real-world WSN. One synthetic graph is a randomly generated Erdős–Rényi (ER) graph with 200 nodes and 0.2 edge probability \cite{erdos1959graph}. The real-world WSN is collected from 54 sensors deployed in the Intel Berkeley Research lab \footnote{ \url{http://db.csail.mit.edu/labdata/labdata.html}.}. 

\textbf{Communication model and sample generating} The number of worms are select from ${2, 4, 8}$. For each node $v$, the threshold value of worm $k$, $\theta_v^k$, is generated uniformly random from $[0, 1]$. The weight on each edge $(u,v)$ of worm $k$ is set as $w_{(u,v)}^k = \frac{1}{|N_v| + k}$. For each graph, we generate a sample pool $D = \{(\mathcal{I}_0, \mathcal{I}_N)^i\}_i^{1000}$ with $N/2$ seeds and each seed is selected uniformly at random. Each selected seed node is assigned to worm uniformly at random. Given $\mathcal{I}_0$, $\mathcal{I}_N)$ is generated follow the communication model.

\textbf{Baseline method.} We consider two baseline algorithms: RegularNN and Support Vector Machine (SVM). RegularNN is a real-valued deep neural network model with $4\cdot N$ layers. The activation functions are set as a combination of linear, ReLU, and sigmoid. We aim to compare ComplexNN, which explicitly simulates the diffusion process, with the regular designed neural network. Since the proposed problem is a binary classification task, we implement the SVM algorithm. Besides, the random method with a random prediction is also considered. 

\textbf{Implementation} The size of the training set is selected from $\{200, 400, 600\}$ and the testing size is $400$. During each simulation, the train and test samples are randomly selected from the sample pool. For each method, we use F1 score, precision, recall, and accuracy to evaluate the performance. Besides, the number of simulations for each setting is set to 5, and we report the average performance together with the standard deviation.

\begin{table}[]
\centering
\caption{\textbf{Impact of the different number of worms. } Each simulation is implemented with 600 training samples, 400 testing samples, and 100 seeds.}
\label{tab:my-table}
\vspace{2mm}
\resizebox{0.85\textwidth}{!}{%
\begin{tabular}{@{}ccccc@{}}
Number of worms & F1                 & Precision          & Recall             & Accuracy           \\ \midrule
2               & 0.717\tiny(3.E-02) & 0.657\tiny(1.E-02) & 0.948\tiny(3.E-02) & 0.803\tiny(2.E-02) \\
4               & 0.469\tiny(6.E-03) & 0.419\tiny(6.E-03) & 0.629\tiny(6.E-03) & 0.727\tiny(3.E-03) \\
8               & 0.277\tiny(4.E-03) & 0.206\tiny(5.E-03) & 0.313\tiny(5.E-03) & 0.656\tiny(3.E-03)
\end{tabular}%
}
\end{table}

\begin{table}[]
\centering
\caption{\textbf{Impace of the different number of seeds.} Each simulation is implemented with 600 training samples, 400 testing samples, and 4 worms.}
\label{tab:seeds}
\vspace{2mm}
\resizebox{0.85\textwidth}{!}{%
\begin{tabular}{@{}cllll@{}}
Number of seeds & \multicolumn{1}{c}{F1 score}           & \multicolumn{1}{c}{Precision}          & \multicolumn{1}{c}{Recall}             & \multicolumn{1}{c}{Accuracy}           \\ \midrule
50              & 0.332\tiny(4.E-03)                     & 0.277\tiny(4.E-03)                     & 0.579\tiny(7.E-03)                     & 0.676\tiny(4.E-03)                     \\
100             & \multicolumn{1}{c}{0.469\tiny(6.E-03)} & \multicolumn{1}{c}{0.419\tiny(6.E-03)} & \multicolumn{1}{c}{0.629\tiny(6.E-03)} & \multicolumn{1}{c}{0.727\tiny(3.E-03)} \\
150             & 0.549\tiny(5.E-03)                     & 0.524\tiny(4.E-03)                     & 0.626\tiny(4.E-03)                     & 0.733\tiny(3.E-03)                    
\end{tabular}%
}
\end{table}

\subsection{Observations}

Table \ref{tab:main} shows the main results. Each cell includes the average value and its corresponding standard deviation. We can observe that our method outperforms other methods. Compared to the RegularNN, the proposed method are able to improve the prediction errors. And this shows the advantage of using an explicit model. Besides, it shows that the performance of each algorithm is increased given a larger graph.

\textbf{Impact of number of worms} Table \ref{tab:my-table} gives the simulation results for our proposed method under the different numbers of worms. We can observe that with a lower number of worms, the performance is improved. One plausible reason is that the number of worms determines the size of the number of input and output features for our model. And it meets the nature of neural network training properties.

\textbf{Imbalanced class distribution} The simulation results for our proposed method under the different number of seeds are given in Table \ref{tab:my-table}. We see that the performance is increased when more seeds are provided. Since our model is a binary classifier, a small number of seeds leads to an imbalance class distribution. It is an inherent issue in predicting the node status in that the majority of the nodes tend to be inactive.

\section{Conclusion}
In this paper, we formulate a propagation model for multiple worms and propose a complex-valued neural network model. In particular, our designed complex-valued neural network model simulates the propagation process exactly. Overall our method outperforms classic supervised learning methods. However, it is not sufficiently stable for different data within the different numbers of worms and seeds. And it is possible that there exists an advanced representation strategy for the neural network design that can achieve a better performance, which is one direction for our future work. In addition, encouraged by the possibility of using a complex-valued neural network to simulate the worms' propagation models, we believe it would be interesting to apply our method to the traditional epidemic model.

%
%
%
\bibliographystyle{splncs04}
\bibliography{mybibliography}

%




\end{document}